%% file: unsup_faces.tex
\documentclass[10pt, conference, compsocconf]{IEEEtran}

\ifCLASSINFOpdf
  \usepackage[pdftex]{graphicx}
\fi
\usepackage[cmex10]{amsmath}
\usepackage{float}

\usepackage{amssymb}  
\usepackage{multirow}
\usepackage{tabularx, booktabs}
\newcolumntype{C}{>{\centering\arraybackslash}X}
\usepackage[T1]{fontenc}

\begin{document}
\title{Unsupervised Learning of Face Representations}

\author{
	\IEEEauthorblockN{
		Samyak Datta\IEEEauthorrefmark{1}\IEEEauthorrefmark{2},
		Gaurav Sharma\IEEEauthorrefmark{3},
		C.V. Jawahar\IEEEauthorrefmark{2}
	}
	\IEEEauthorblockA{
		\IEEEauthorrefmark{1}Georgia Institute of Technology,
		\IEEEauthorrefmark{2}CVIT, IIIT Hyderabad,
		\IEEEauthorrefmark{3}IIT Kanpur
	}
}
\maketitle

\newcommand{\etal}{\textit{et al}. }
\newcommand\blfootnote[1]{%
	\begingroup
	\renewcommand\thefootnote{}\footnote{#1}%
	\addtocounter{footnote}{-1}%
	\endgroup
}

\begin{abstract}
We present an approach for unsupervised training of CNNs in order to learn discriminative face
representations. We mine supervised training data by noting that multiple faces in the same video
frame must belong to different persons and the same face tracked across multiple frames must belong
to the same person. We obtain millions of face pairs from hundreds of videos without using any
manual supervision. Although faces extracted from videos have a lower spatial resolution than those
which are available as part of standard supervised face datasets such as LFW and CASIA-WebFace, the
former represent a much more realistic setting, e.g.\ in surveillance scenarios where most of the
faces detected are very small. We train our CNNs with the relatively low resolution faces extracted
from video frames collected, and achieve a higher verification accuracy on the benchmark LFW
dataset cf.\ hand-crafted features such as LBPs, and even surpasses the performance of
state-of-the-art deep networks such as VGG-Face, when they are made to work with low resolution
input images.

\end{abstract}


\IEEEpeerreviewmaketitle

\section{Introduction}
\blfootnote{$^*$Work done while Samyak Datta was with CVIT, IIIT Hyderabad}
Recent advances in deep convolutional neural networks, e.g.\ Krizhevsky \etal
\cite{krizhevskyNIPS2012}, coupled with availability of large annotated datasets, have led to
impressive results on supervised image classification. While the annotated datasets available are of
the order of millions of images, annotation beyond a point is infeasible, especially to the scale of
billions of images and videos available on the Internet currently. To leverage this vast amount of
visual data, albeit without any annotations, the community is starting to look towards unsupervised
learning methods for learning generic image representations \cite{doersch2015unsupervised,
	goroshin2015unsupervised, le2013building, srivastava2012multimodal}. The general idea is to obtain
some form of weakly supervised data and then train a system to make predictions of these
meta-classes from the originally unannotated images, e.g.\ Huang \etal \cite{huang2016unsupervised}
use a discriminative clustering method to mine attributes and then learn a CNN based similarity
to capture the presence of attributes in the images while Wang \etal \cite{wang2015unsupervised}
propose to derive similar and dissimilar patches via motion information and then learn a CNN based
similarity, conceptualy close to the former method. These methods have shown good promise for the
general task of image representation learning. 
\begin{figure}
	\begin{center}
		\includegraphics[width=\columnwidth,trim=0 10 280 0,clip]{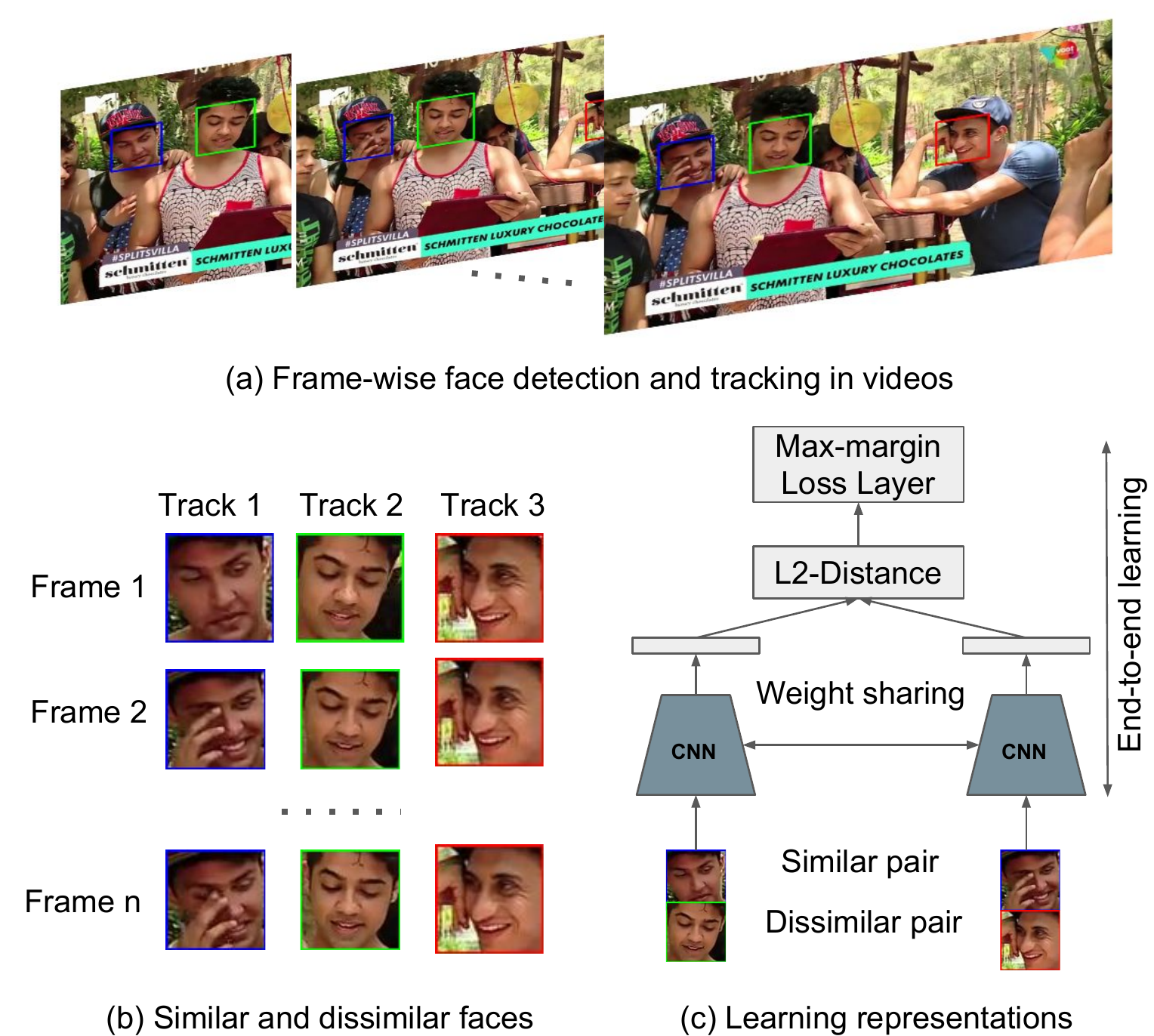}
	\end{center}
	\vspace{-1em}
	\caption{Overview of our approach. (a) Given a video, we perform frame-wise face detection followed by the temporal tracking of those faces across different frames. (b) Based on the observation that multiple faces in the same frame must belong to different persons and the same face tracked across multiple frames must belong to the same person, we generate a set of similar and dissimilar face pairs. (c) We use these face pairs to train a Siamese network using max-margin loss that learns discriminative face representations.}
	\label{fig:overview}
	\vspace{-1.5em}
\end{figure}

In the present paper, we focus on a particular but very important subset of images, that of human
faces, and investigate unsupervised learning of representations for the same. We study if
unsupervised learning can compete with hand-crafted features as well as with supervised learning
based methods, for the specific and limited case of human face images. Given the availability of
large amount of human centered data on the Internet, we investigate whether having a large corpus of such
videos with challenging appearance variations makes it possible to learn a meaningful
face representation which can be used out-of-the-box for face verification in the wild. Also, do the representations
improve when they are fine-tuned on the current, application and domain conditioned set of faces? In
particular, since getting very high resolution data is impractical so far, in real life scenarios
like surveillance etc., we focus on lower resolution face images. We show that using a simple
observation that faces in the same frame have to be of different persons (barring rare exceptions of
reflections etc.) and those which are associated via tracking across different frames are of the
same person, we train a CNN based similarity function. We then used this pre-trained network to
obtain the representations of novel features for face verification on novel datasets. We also
finetune these representations on the given task and show how the performance changes.  As a
summary, the main contributions of our paper are as follows.
(i) We collect a large-scale dataset of 5M face pairs with similar and dissimilar labels by
tracking YouTube videos. The dataset is shown to have faces in a wide variety of scale, pose,
illumination, expression and occlusion conditions, thereby  making it challenging source of data for
facial analysis research.
(ii) We present a method to learn discriminative, deep CNN based facial representations using the
above dataset. In contrast with other state-of-the-art facial representations, we do not use even a
single identity label for learning our representations. Rather, we rely on weak annotations of
similarity and dissimilarity of faces to train our network. 
(iii) We perform extensive empirical studies and demonstrate the effectiveness of the
representations that we learn by comparing them with hand-crafted features (LBP) as well as
state-of-the-art VGG-Face descriptor in the domain of low-resolution input images. We also do
ablation studies and report the importance of the different parametric choices involved.

\section{Related Work}
Supervised image classification has seen rapid progress recently, starting from the seminal works of
Krizhevsky \etal \cite{krizhevskyNIPS2012}. While the progress has been less stellar for the
unsupervised counterparts, the field has been active in studying such methods as well
\cite{bengio2013representation, doersch2015unsupervised,
	hinton2006reducing, le2013building,
	lee2009convolutional}.
Particularly in computer vision, there have been numerous attempts at learning representations
using, e.g., videos \cite{le2011learning,
	srivastava2015unsupervised, zou2012deep, goroshin2015unsupervised,
	mobahi2009deep, stavens2010unsupervised}
As some representative works, Wang \etal \cite{wang2015unsupervised} mine thousands of unlabelled
videos to track objects in order to learn effective representations. Masi \etal \cite{masi2016we}
uses low-level motion cues to perform motion-based segmentation of objects in videos which are then
used as ground truth for training CNNs.  Liang \etal \cite{liang2015towards} uses an Image-Net
pre-trained CNN as initialization followed by semi-supervised learning in videos for object
detection. 

The progress, in the specific domain of human faces, has also been dominated by supervised face
recognition methods based on deep neural networks. The DeepID series of networks
\cite{sun2015deeply} were trained using two types of supervisory
signals -- a Softmax loss that learns to separate inter-personal variations and a pairwise
verification loss that learns intra-personal variations.  These networks were trained using identity
labeled images from the CelebFaces+ \cite{liu2015faceattributes} dataset. The training using Softmax loss was done
using a randomly sampled set of $8192$ identities and $200$k pairs from the remaining $1985$
identities were used for learning the verification model.  Similarly, \cite{parkhi2015deep} poses
face recognition as an $N$-way classification problem where $N = 2600$. They collect a identity
labelled dataset  of $2.6$M face images with the help of popular image search engines. The authors
then train their networks using a Softmax loss to learn identity-discriminative face
representations. On the other hand, \cite{schroff2015facenet} and \cite{taigman2014deepface} use
weak annotations in terms of either triplets or (non-)matching pairs to train deep network to learn
representations that discriminate between similar and dissimilar face pairs. While
\cite{taigman2014deepface} uses 4 million face images belonging to 4k identities, the dataset
collected in \cite{schroff2015facenet} consists of 200 million training images. Although our work
also makes use of weak annotations in the form of (dis)similar pairs, unlike all the methods
discussed here, we do not use a single identity label in the process of generating our training
pairs.

Curating large-scale, identity labeled face datasets poses challenges in scaling. As a testimony to
this fact, the biggest supervised face datasets have been collated by organizations such as Google
\cite{schroff2015facenet} and Facebook \cite{taigman2014deepface} and are not accessible to the
research community. This raises a very pertinent question -- is it really necessary to use millions
of identity labeled face images to learn discriminative representations?  Masi \etal \cite{masi2016we} 
tackle this problem by augmenting labelled datasets by synthesizing face images in challenging appearance 
variations. A complimentary approach to deal with the same challenge is to develop methods and systems 
to leverage the large amounts of unsupervised video data, freely available on the internet -- we
explore this approach here. Both \cite{wang2015unsupervised} and \cite{cinbisICCV2011} and used
motion information to obtain supervision of similar and dissimilar patches (based on objects or
faces, respectively. Since, like us, Cinbis \etal \cite{cinbisICCV2011} work with faces, we point
out the following differences between our work and them.  While their scope of application is quite
narrow, i.e.\ learning cast-specific metrics for TV videos, we are interested in whether we can learn
meaningful face representations without having to rely on supervision in terms of face identities.
We work with end to end pipelines for learning representation, while they have hand designed
features. We work at a significantly larger scale; while they had $3$ videos (episodes) with
$\sim650$ manually annotated tracks, we work with $850$ videos with $1904$ automatically formed
tracks giving a total of $5\times 10^6$ mined similar and dissimilar face pairs. Similarly, the
number of identities they had were 8, while here we have order of hundreds. Also, they acknowledge
(Sec.~3.2) that their set of dissimilar pairs is heavily biased towards a few pairs of actors who
co-appear frequently, while we have taken specific measures to mitigate such a scenario
by doing cross-pairings between faces collected from widely different genres of videos.
Finally, they worked with high quality broadcast videos whereas we work with unconstrained YouTube
videos, and study low resolution face analysis and, while they used significant manual annotations,
large majority of the steps are automatic in our case and here our method is scalable.

Recently, there also has been an interest towards learning face representations using low
resolutution images. Hermann et al.~\cite{herrmann2016low_1} train networks on
input image sizes of $32 \times 32$ by combining face images from multiple data sources including
surveillance videos--which are a very large and relevant area for face identification applications.
All images in their dataset, however, have been manually annotated with identity labels, posing
scalability challenges. We aim to learn low resolution face representations without manual identity
level annotations.

\section{Overview}
Our goal is to train deep convolutional neural networks to learn face representations using videos
downloaded from the internet. High-performing, state-of-the-art systems rely on massive amounts of
high-resolution, identity-labeled face images for learning discriminative face representations.
Table \ref{tab:datasetFaceSizes} shows the statistics of different face datasets being used, most of
the existing datasets use faces of sizes upwards of $\sim 120\times 120$. As an example, VGG-Face
network \cite{parkhi2015deep} was trained on $2.6$ million face images annotated with identity
labels from a set of $N = 2600$ celebrity identities with an input image size of $224 \times 224$.
They posed face recognition as an $N$-way classification problem and trained their network
architecture using a standard classification (Softmax) loss. The $\ell_2$-normalized output of the
last fully connected layer was then used as a face descriptor which was further finetuned on the LFW
training set. 

We work with a large set of internet videos, i.e.\ they do not provide us with any strong
supervision in the form of identity labels for faces. This rules out the possibility of using a
set-up similar to VGG-Face and learning identity-wise discriminative representations. However, we
exploit the temporal supervision that is implicit in videos to generate weak annotations in terms of
similar and dissimilar (wrt.\  identity) face pairs. Figure \ref{fig:overview} gives an illustration
of our overall approach, which we explain in detail below. Instead of using identity labeled face
images, we train our CNNs by using the generated pairs of faces. The visual representations that we
learn embeds similar face pairs closer in the feature space than dissimilar faces using a max-margin
loss function that learns to separate similar and dissimilar pairs by a specified margin. We now
give the details of all the different steps involved.

\begin{table}
	\begin{center}
		\begin{tabularx}{\columnwidth}{@{}c*{5}{C}c@{}}
			\toprule
			\multicolumn{1}{c}{\multirow{2}{*}{Dataset}}  & \multicolumn{1}{c}{\multirow{2}{*}{Id}} & \multicolumn{1}{c}{\multirow{2}{*}{\# Faces}} & \multicolumn{2}{c}{Face size} & \multicolumn{1}{c}{\multirow{2}{*}{Ratio}} \\
			\multicolumn{1}{c}{}  & \multicolumn{1}{c}{} & \multicolumn{1}{c}{} & \multicolumn{1}{c}{mean}  & \multicolumn{1}{c}{median} & \multicolumn{1}{c}{} \\
			\midrule
			LFW 		\cite{huang2007labeled} 	& \checkmark & 	13k	& 113   &  113	& 1.7$\times$ \\
			TSFT		\cite{sharmaCVPRW2015}		& \checkmark & 	32k	& 121	&  109  & 1.8$\times$ \\
			VGG-Face  	\cite{parkhi2015deep} 		& \checkmark & 	2.6M	& 125   &  108  & 1.9$\times$ \\
			CelebA 		\cite{liu2015faceattributes}	& $\times$   & 	220k	& 152	&  145 	& 2.3$\times$ \\
			Ours   						& $\times$   & 	1.36M	& 67    &  56 	& ref. \\
			\bottomrule
		\end{tabularx}
		\caption{Face sizes (mean and median) across different datasets -- both supervised and unsupervised, i.e.\ with and
			without identity labels resp. (``Id'' column). The ``Ratio'' column gives the relative mean face size with respect to the mean face size in our dataset. } 
		\label{tab:datasetFaceSizes}
	\end{center}
	\vspace{-1.5em}
\end{table}

\begin{figure}
	\begin{center}
		\includegraphics[width=\columnwidth,trim=0 0 280 0,clip]{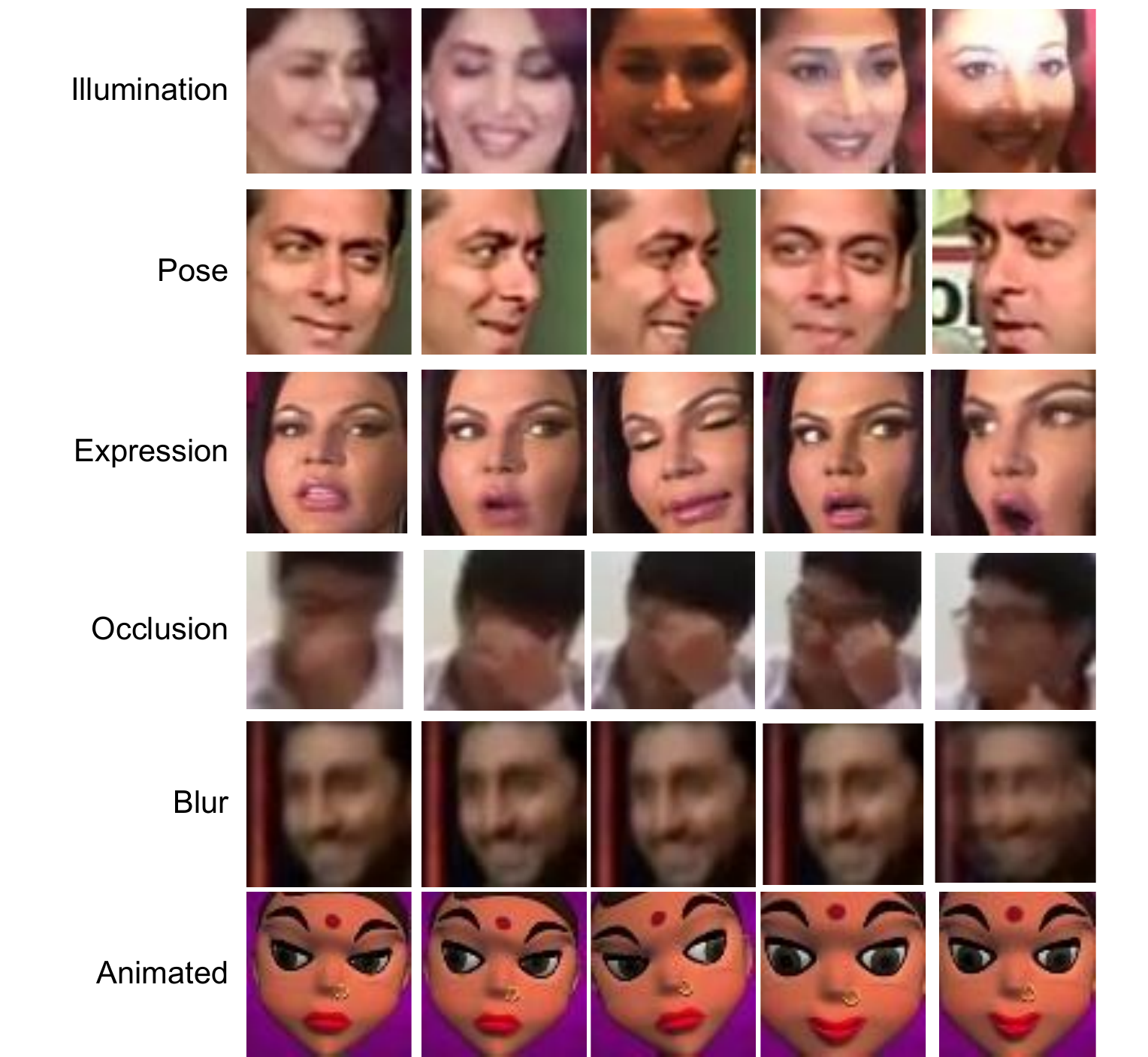}
	\end{center}
	\vspace{-1.5em}
	\caption{Some samples showing the rich amount of variations in terms of pose, occlusion, illumination, expression etc. in the dataset that we have captured}
	\label{fig:unsupDatasetVariations}
	\vspace{-1.5em}
\end{figure}

\section{Mining Face Pairs from Videos}

We now describe the data collection process that we used to mine face pairs via face detection and
tracking in videos. A key observation is that all pairs of faces that are detected in a single video
frame must belong to different identities (barring rare exceptions such as reflections in mirrors)
and hence contribute to the set of dissimilar face pairs. Similarly, by tracking a face across
multiple frames we can obtain face images belonging to the same identity which helps us generate our
set of similar face pairs. By processing thousands of videos in this fashion, we were able to
generate a dataset of more than 5 million face pairs as follows.

\subsection{YouTube Videos}
To construct our dataset we downloaded videos from YouTube - a popular video sharing site. Since our
aim was to detect faces in order to generate a set of face pairs, we chose the genres of videos in
such a manner so as to maximize the number of faces appearing in each video frame-- genres which
have a lot of people appearing in them and at the same time instants, and with high number of faces
appearing in any given video frame. Keeping this in mind, we looked at videos of news debate shows,
TV series where multiple actors are part of a majority of the scenes, discussion panels with several
panelists, reality shows where participation is usually in the form of teams/groups and celebrity
interview shows. We manually created a list of search keywords on the basis of the above criteria for genre selection and
downloaded $850$ videos from YouTube.

Given a video, we then wanted to detect faces and subsequently track the faces over time to generate
the desired set of similar and dissimilar face pairs.

\subsection{Face Detection}
We used the implementation of the Viola-Jones (VJ) face detector \cite{viola2001rapid} that is available as part of the
OpenCV library  for detecting faces. Every 10th frame of a given video was sampled and a
combination of face detectors trained on frontal and profile faces was applied to detect faces that
appear in the frames. The face detectors were configured to output only high-confidence detections
at the cost of missing out on some faces. There were some cases of false positive detections which
were removed manually (approximately 10-12\% of the total face detections were false positives).

\begin{figure}
	\begin{center}
		\includegraphics[width=\linewidth,trim=10 10 10 10,clip]{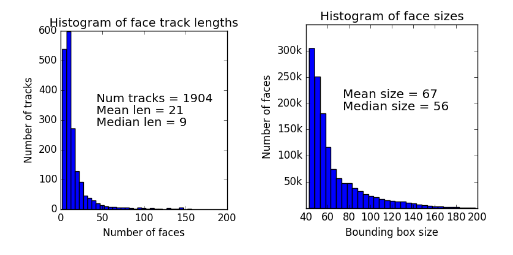}
	\end{center}
	\vspace{-1.5em}
	\caption{Histogram of track lengths and face sizes (square bounding box) in the proposed dataset collected in an unsupervised fashion from videos. The average face size and track length is 67 pixels and 21 faces respectively.}
	\label{fig:datasetStats}
	\vspace{-1.5em}
\end{figure}

\subsection{Face Tracking}
Given the location of detected faces (if any) in each of the sampled video frames, the next step was 
to generate face tracks. Face tracking was done using a tracking-by-detection framework with a
temporal association of face bounding boxes across frames. For each detected face in a given frame,
we checked if there was an overlap between its bounding box with that of any face in the previously
sampled frames. If there was an overlap, the current face was added to the existing track (if there
are multiple overlaps, we added it to the track corresponding to the face with the maximum overlap).
Otherwise, a new face track was started for the given face detection. A face track was marked to have
ended when no new bounding boxes were added for $5$ consecutive sampled frames. As a post-processing
step, all face tracks with less than $5$ faces were discarded. This rather simple and conservative
tracking strategy ensured that we got only high quality similar face pairs, which is eventually what
the tracking was supposed to achieve.

\subsection{Dataset Statistics}
After processing all the $850$ videos with a total running time of about $70$ hours, we detected a
total of $1.36$M faces and generated $1904$ face tracks. The distribution of face track length
(number of faces in a track) as well as the distribution of face sizes (square bounding box as
detected by the VJ detector) are shown in Figure \ref{fig:datasetStats}. Given the faces detected in
a single frame as well as the face tracks, we generated a set of more than $5$ million similar
and dissimilar face pairs. Table \ref{tab:unsupDatasetStats} shows the statistics of the constructed
face pairs dataset. A comparison of the face sizes in our dataset in Table
\ref{tab:datasetFaceSizes} shows that faces detected from video frames are of much lower spatial
resolution than those available as part of identity-labeled face datasets such as LFW, CelebA etc.
The mean face size as detected by the VJ detector in our dataset is $67$ pixels whereas it is $113$
pixels for LFW and goes up to $152$ pixels in the case of CelebA. This fact plays an important role
in designing our CNN architecture which is discussed in Section \ref{sec:learning}.  Qualitatively,
as shown in Figure \ref{fig:unsupDatasetVariations}, the dataset is able to capture a rich amount of
variations in challenging conditions of pose, illumination, expression, resolution and occlusion
etc., making it an apt source for learning face representations.

\begin{table}
	\begin{center}
		\begin{tabularx}{0.50\columnwidth}{@{}c*{1}{C}c@{}}
			\toprule
			\# videos & 850 \\
			duration & 70 h\\
			\# faces & 1.36 M \\
			\# face tracks & 1904 \\
			\# similar pairs & 2.5 M \\
			\# dissimilar pairs  & 2.5 M \\
			\toprule
			\# total pairs & 5 M \\
			\toprule
		\end{tabularx}
		\label{tab:unsupDatasetStats}
		\caption{Statistics of the proposed unsupervised face pairs dataset that was collected from videos.}
	\end{center}
	\vspace{-1.5em}
\end{table}

\section{Learning Face Representations}
\label{sec:learning}
The previous section described how we generated similar and dissimilar face pairs from unlabeled
videos. In this section, we discuss our framework that uses those millions of face pairs and learns
discriminative face representations.

\subsection{Network Architecture}
We designed a Siamese network which consists of two base networks which share parameters and whose
architectures (convolutional layers) are similar to the VGG-Face network \cite{parkhi2015deep}. The
VGG-Face network was trained using images of size $224 \times 224$. As shown in Figure
\ref{fig:datasetStats} and Table \ref{tab:datasetFaceSizes}, the face sizes in our dataset are
constrained by the video quality and are much smaller than $224 \times 224$. Therefore, we trained
our networks for input resolutions of $64 \times 64$ and $128 \times 128$. As a consequence of this
change in resolution of input images, the sizes of the FC layers were also changed appropriately. 
Instead of 4096-$d$, we have 1024-$d$ FC layers, and cf.\ the architecture of \cite{parkhi2015deep}
which consisted of 38M parameters, our networks have 17 million and 24 million parameters, for the
$64$ and $128$ sized networks, respectively. The final output (of the last FC layer) of both our
nets is a 1024-$d$ feature vector that is the learned representation of the face in the feature
space.

\subsection{Max-margin Loss Function}
If $\mathbf{x}$ is an input to the CNN, let $\phi(\mathbf{x})$ be the output of the last FC layer.
The function $\phi(.)$ is parameterized by the weights and biases of the CNN. We define the distance
between two faces $\mathbf{x_1}$ and $\mathbf{x_2}$ in the learned representation space as the
square of the L2 distance between their descriptors, i.e.\
\begin{equation}
\label{eq:L2distance}
D^2(\mathbf{x_1}, \mathbf{x_2}) = \| \phi (\mathbf{x_1}) - \phi (\mathbf{x_2}) \|^2_2
\end{equation}

Our goal was to learn visual representations such that similar face pairs are closer together in the
representation space and dissimilar face pairs are far apart. This objective guided our choice of
the loss function that we used for training. Formally, let $\mathcal{T} = \{ (\mathbf{x_i},
\mathbf{x_j}, y_{ij}) \}$ be the dataset where $y_{ij} = 1$ if $\mathbf{x_i}$ and $\mathbf{x_j}$ are
similar face pairs and $y_{ij} = -1$ otherwise. Then, the loss function can be defined as follows,
\begin{equation}
\label{eq:maxMarginLoss}
L(\mathcal{T}) = \sum_{\mathcal{T}} \max \left\{ 0, m - y_{ij}(b - D^2(\mathbf{x}_i, \mathbf{x}_j))
\right\}.
\end{equation}
Where, $D(\mathbf{x_i}, \mathbf{x_j})$ is the distance as defined in Eq.~\ref{eq:L2distance}.
Minimization of this margin-maximizing loss encourages the distance between pairs of faces of same
(different) person to be less (greater) than the bias $b$ by a margin of $m$. 

\subsection{Fine-tuning Using Supervised Metric Learning}
Supervised fine-tuning of descriptors using the target dataset has been shown to be very effective
for faces \cite{simonyan2013fisher}. Using the max-margin loss as described in
Eq.~\ref{eq:maxMarginLoss}, the CNN learns discriminative representations for faces by training over
data from the (unsupervised) face pairs dataset. Since we wished to report pair matching (face
verification) accuracies on the benchmark LFW dataset, we used training pairs from LFW to fine-tune
our descriptors using metric learning. This was done by learning low dimensional projections with a
discriminative objective function. Therefore, the fine-tuning served two purposes - (a) it reduced
the dimensionality of the learned representations, making it suitable for large datasets, and
(b) further enhanced the discrimination capability of the features upon projection.

Formally, the aim was to learn a linear projection $W \in \mathbb{R}^{p \times d}$, $p \ll d$ which
projects the representations learned by our network $ \phi(\textbf{x}) \in \mathbb{R}^d $ to low
dimensional projections $W\phi(\textbf{x}) \in \mathbb{R}^p$. This was done such that the square of
the Euclidean distance between faces $i$ and $j$ in the projected space, given by the equation
\begin{equation}
\label{eq:fineTuningDistance}
D^2_W(\phi_i, \phi_j) = \| W\phi_i - W\phi_j \|^2_2,
\end{equation}
is smaller than a learned threshold $b \in \mathbb{R}$ by a (fixed) margin of $m$, if $i$ and $j$
are of the same person, and larger otherwise. 

The objective is similar to the one which we used to train our deep network. Such fine-tuning using
metric learning is equivalent to supervised learning of another FC layer on top of the unsupervised
representations learned by our CNN. There is one difference -- while training the deep network, both
the bias and the margin were fixed whereas while fine-tuning, only the margin is fixed and the bias
is a parameter which is learned.

\begin{figure}
	\begin{center}
		\includegraphics[width=\columnwidth,trim=23 20 490 10,clip]{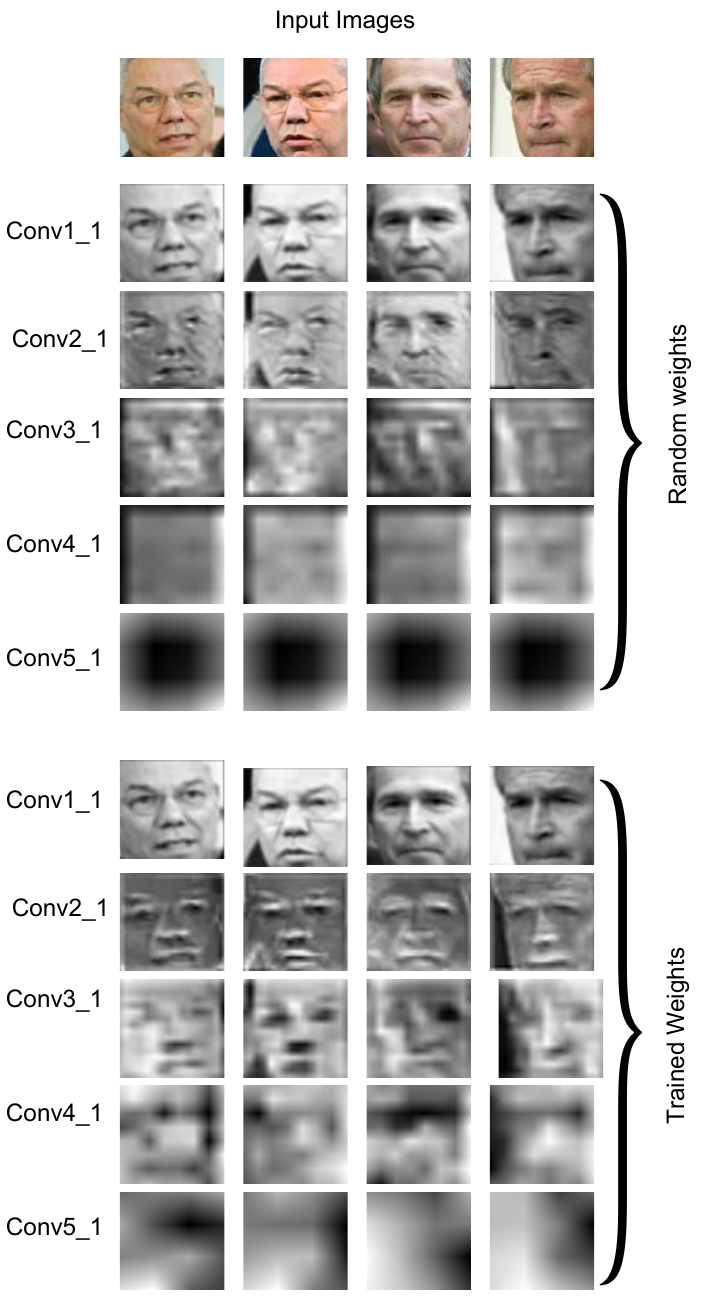}
		\vspace{-2em}
		\caption{Visualization of the activations of the 1st filter for 5 Conv. layers at different depths for 4 input images from LFW. A qualitative comparison between our trained network and a network with random weights shows the discriminative abilities of the weights learned by our network especially at the deeper layers.}
		\label{fig:activationMaps}
	\end{center}
	\vspace{-1.5em}
\end{figure}

\begin{table*}
	
	\begin{center}
		\begin{tabularx}{1.0\textwidth}{@{}lc*{4}{C}c@{}}
			\toprule
			
			\multicolumn{1}{c}{\multirow{2}{*}{\hspace{12em}}}  & \multicolumn{2}{c}{No fine-tuning
				done (accuracy | EER | AUC)} & \multicolumn{2}{c}{Fine-tuned on LFW (accuracy | EER |
				AUC)} \\ 
			
			\addlinespace
			
			\multicolumn{1}{c}{Descriptors $\downarrow$ \ \  Image sizes $\rightarrow$} &
			\multicolumn{1}{c}{$64 \times 64$} & \multicolumn{1}{c}{$128 \times 128$} &
			\multicolumn{1}{c}{$64 \times 64$} & \multicolumn{1}{c}{$128 \times 128$} \\
			\midrule
			
			LBP  &  64.60 $\vert$ 35.40 $\vert$ 70.79 &  64.60 $\vert$ 35.39
			$\vert$ 70.39& 70.18 $\vert$ 29.82 $\vert$ 78.14 & 72.44 $\vert$ 27.56 $\vert$ 79.82  \\
			\addlinespace
			
			VGG-Face (Random) & 60.54 $\vert$ 39.46 $\vert$ 64.97 & 61.55 $\vert$
			38.45 $\vert$ 65.72   & 65.02 $\vert$ 34.98 $\vert$ 70.65 & 65.34 $\vert$ 34.66 $\vert$ 71.64  \\ 
			\addlinespace
			
			VGG-Face (Supervised)  & 62.38 $\vert$ 37.62 $\vert$ 67.21  & 62.55 $\vert$ 37.45
			$\vert$ 67.43  & 66.00 $\vert$ 34.00 $\vert$ 72.08 & 66.34 $\vert$ 33.67 $\vert$ 73.03 \\
			\addlinespace
			
			Proposed face representations & 71.48 $\vert$ 28.53 $\vert$ 78.78  &  71.48 $\vert$ 28.51
			$\vert$ 78.40  & 73.22 $\vert$ 26.78 $\vert$ 80.57  & 72.20 $\vert$ 27.79 $\vert$ 80.29  \\
			
			\bottomrule
			
		\end{tabularx}
		\caption{Face verification accuracies for different descriptors across
			different input image sizes: (i) hand-crafted features -- \emph{LBP}, (ii) a network (with the same
			architecture as ours) initialized with random weights -- \emph{VGG-Face (Random)}, (iii) the
			VGG-Face network \cite{parkhi2015deep} pre-trained network made to operate in a
			low-resolution setting -- \emph{VGG-Face (Supervised)}, and (iv) peoposed method. 
		} 
		\vspace{-.9em} \label{tab:accuracy}
	\end{center}
	\vspace{-3em}
\end{table*}

\section{Experiments}
We now present the experiments we did to validate our method. We first give the training details,
followed by results with unsupervised representations learnt by the networks. We then give the
results when the unsupervised representations are fine-tuned on the current task. We also give
ablation studies, showing the importance of difference choices.

\subsection{Training Details}

The training dataset comprises of $2.5$M similar and dissimilar face pairs each. Apart from pairing
faces that were detected in the same video frame, we increased the number of dissimilar face pairs
by taking arbitrary subsets of faces detected from widely different genres of videos and pairing
them up. For example, the video of a TV reality show based in another country is highly likely to
have a mutually exclusive set (in terms of identities) of actors/participants than a popular
Hollywood TV series. We found such forms of cross-pairings to be an effective way to increase the
size of the dataset. 
\vspace{0.8em}\\
\textbf{Unsupervised Training Implementation.}
We trained two networks with different input resolutions -- $64 \times 64$ and $128 \times 128$
using backpropoation and SGD with a batch-size of $32$ image pairs and a learning rate of $0.01$. For
regularization, we set the weight-decay parameter to $0.0005$ and also do Batch-Normalization after
every convolutional and FC layer. The bias and margin of the max-margin loss ($b$ and $m$ in Eq.
\ref{eq:maxMarginLoss}) were set to $1.0$ and $0.5$ respectively. All implementations related to
training of the CNNs were done using Torch and the networks were
trained on $12$ GB NVIDIA GeForce TitanX GPUs.

For validation, we kept aside a set of $6400$ similar and dissimilar face pairs each as our
validation set (this was disjoint from our training set). We monitored the training of our CNN using
face matching accuracies on the validation set with the bias of the loss function as a threshold. A
(dis)similar face pair in the val set is said to be correctly matched if the distance between their
descriptors is (greater) less than the threshold.
\vspace{0.8em}\\
\textbf{Hard-mining.}
We trained the networks until the validation accuracy began to saturate and then performed
hard-mining of the training pairs. During hard-mining, we computed the distances between the learned
descriptors of all face pairs in the training set. Those (dis)similar pairs whose distances are
greater (lesser) than $b (+)- m$ were classified as ``hard pairs'' which meant that they would
accrue a loss as per our max-margin loss module. After hard-mining, training was resumed only on the
``hard  pairs''. This enabled the CNN to learn more robust representations.

The hard-mining process was performed after $28$k iterations and $50$k iterations for the $64 \times
64$ and the $128 \times 128$ net respectively. Post hard-mining, $28.07\%$ and $16.18\%$ of the
original set of $5$M face pairs were classified as ``hard'' for the $64 \times 64$ and $128 \times
128$ net respectively. Both networks were subsequently trained on the reduced set of hard-mined
pairs for $3$ more epochs.
\vspace{0.8em}\\
\textbf{Data Augmentation.}
Similar to \cite{parkhi2015deep}, we performed data augmentation during training. We scaled the
input images to a slightly larger size by keeping the aspect ratio same and then proceeded to take
crops of the desired input size from all the $4$ corners and the center resulting in $5$ different
images. We also performed a horizontal flip for each of the face crops to get a total of $10$ data
augmented versions of a single image. These augmentations were applied randomly and independently to
each image of any given face pair during training. In total, we trained the $64 \times 64$ and $128
\times 128$ networks for a total of 150k iterations and 125k iterations respectively (for a batch size of 32 and dataset size of 5M face pairs, this corresponds to 0.8 and 1 epoch respectively.). 

\subsection{Results with Unsupervised Representations}
We refer to the representations learned by our network after training on the face pairs dataset as
``unsupervised representations''. In this section, we demonstrate the effectiveness of the
unsupervised representations using both qualitative and quantitative experiments. 
\vspace{0.8em}\\
\textbf{Qualitative Results.}
We sampled four images from the LFW dataset -- 2 images each belonging to 2 identities and
visualized the activations of 5 convolutional layers at different depths for all the 4 images.
Figure \ref{fig:activationMaps} shows a qualitative comparison between the activations for two
networks: (a) initialized with random weights and (b) trained using unsupervised face pairs data
(both the nets have $64 \times 64$ sized inputs). It can be observed from Figure
\ref{fig:activationMaps} that the random weights CNN has very low discriminative power as evident
from its activations from the deeper convolutional layers (Conv4\_1 and Conv5\_1) which are similar
for all the 4 inputs. On the other hand, our trained network has been able to learn weights which
enable it to discriminate between images. 
\vspace{0.8em}\\
\textbf{LFW Dataset.}
Quantitatively, we report face verification (pair matching) accuracies on the benchmark LFW dataset.
The dataset comprises of $13,233$ face images from $5,749$ identities.  We used the Viola-Jones
detector \cite{viola2001rapid} to detect faces in the LFW images. The
average face size (square bounding box) of the faces is $113$ pixels (Figure
\ref{tab:datasetFaceSizes}). The cropped faces were then resized to the desired input size ($64
\times 64$ or $128 \times 128$). For the purpose of pair-wise face matching, the LFW dataset
provides $6000$ face pairs that are divided into $10$ identity exclusive sets (folds). In addition to
the verification accuracies, we also report the area under the ROC curve (AUC) and the Equal Error
Rate (EER) metric which is defined as the error rate when the false positive and the false negative
rates are equal as per the ROC curve. The EER metric is independent of any threshold. All reported
figures (Table \ref{tab:accuracy}) were averaged across the $10$ folds. 

Another noteworthy fact is that we did not use the aligned version of the
LFW dataset as the images in the aligned dataset are all grayscale and our CNNs were trained on RGB
images. Also, we did not perform any sort of facial alignment (other than face detection and
cropping) on the LFW images. We felt that operating in the space of non-aligned, ``in-the-wild''
face images is a more realistic setting. The same principles are also reflected in our training set
which has been shown to posses a wide variety of challenging imaging conditions (Figure
\ref{fig:unsupDatasetVariations}).

\textbf{Quantitative Results.}
The output of the last FC layer of our network was L2-normalized and treated as the learned
(unsupervised) representation of the input face image. In fact, we took equal-sized crops from all 4
corners and the center of the input image, horizontally flipped each of the 5 crops and took the
average of all the 10 descriptors as the descriptor of the input face image. Our CNN, trained using
unsupervised face pairs of size $64 \times 64$, was able to achieve an accuracy of 71.475\% on the
verification task. For the net with an input size of $128 \times 128$, the accuracy was
71.483\%. We expect the accuracies to go up with the size
of the dataset, with diminishing returns. We observed
this in intermediate results while training our networks.
For example, after training for 10k iterations
(which corresponds to a dataset size of 0.3M image pairs), 
the LFW verification rate is 63.35\%. This increases to 64.65\%,
66.92\% and 68.28\% for 0.64M, 1.28M and 1.6M face pairs.
The final verification accuracy that we get using our entire
dataset of 5M training image pairs is 71.48\%. 
\vspace{0.8em}\\
\textbf{Baseline.} We compared the above accuracies with LBP features (hand-crafted features), a
network with randomly initialized weights and the VGG-Face network that has been trained using
supervised data (Table \ref{tab:accuracy}). For LBP, we used the grayscale variants of the LFW
images and kept the cell size as $16 \times 16$. This led to 928-d and 3712-d LBP descriptors for
$64 \times 64$ and $128 \times 128$ input images respectively. To simulate low-resolution input
setting for the VGG-Face network trained using supervised data \cite{parkhi2015deep} , we first
down-sampled the cropped LFW faces to either $64 \times 64$ or $128 \times 128$ and then up-sampled
them to the expected input size of $224 \times 224$\footnote{We also considered training the VGG-Face net from scratch using low-resolution versions of the training images used in \cite{parkhi2015deep}. However, the images in the training set are provided in the form of web URLs and \textasciitilde	25\% of the image URLs were unavailable.}. 

Our network trained on unsupervised data was consistently able to outperform all the others by a significant margin. In comparison to the 71.475\% verification accuracy of our unsupervised network ($64 \times 64$ input size), a network with random weights gave an accuracy of only 60.536\% whereas LBP features gave 64.6\% verification accuracy. The accuracy of the VGG-Face network trained using identity-labeled (supervised) data was only 62.844\%. Our experiments thus demonstrate that the performance of the nets trained on high-resolution (labeled) input faces drops drastically when they are made to operate on a low-resolution setting. These trends are similar to the ones observed in \cite{herrmann2016low_1} where the authors have made similar comparisons. We also plot and compare the ROC curves in Figure \ref{fig:rocCurves} for our network and the baselines (please refer to the Appendix).

\subsection{Results with Fine-tuned Representations}
We also fine-tuned the unsupervised representations learned by our network using training image pairs from the target dataset - LFW. For generating image pairs, we used the ``unrestricted'' setting as mentioned by the dataset providers \cite{huang2007labeled}. The accuracies that we report in Table \ref{tab:accuracy} under the heading ``Supervised (Fine-tuned)'' are all 10-fold cross-validated.

For fine tuning, we initialized the projection matrix $W \in \mathbb{R}^{p \times d}$ using small random values sampled from a zero-mean Gaussian distribution with $\sigma = 0.01$. The margin was set to 0.5 (the bias is a learned parameter). The learning rate was set to $0.01$ and decreased by a factor of 1.2 after every epoch. As expected, there was an increase in the verification accuracies after fine-tuning across all different types of feature descriptors and input image sizes. In the case of our unsupervised representations, as given in Table \ref{tab:accuracy}, the accuracies increased from 71.475\% to 73.220\% for $64 \times 64$ net and from 71.483\% to 72.204\% for the $128 \times 128$ net.

\subsection{Ablation Studies}
We performed some ablation studies, the details of which can be found in the Appendix.


\section{Discussion and Conclusion}
We presented a method to learn discriminative, deep CNN based face representations using a
dataset of 5M face pairs and without using a single identity label. Our net, trained on unsupervised
data, was able to achieve a verification rate of $71.48\%$  on the benchmark LFW dataset with
low-resolution input images of size $64 \times 64$. This perormance is superior to both hand-crafted
features, i.e.\ LBP ($64.6\%$) as well as CNNs ($62.38\%$) trained on supervised data in comparable
low-resolution settings. Further, upon fine-tuning unsupervised representations using max-margin
metric learning on the annotated training images from the LFW dataset, the accuracies for
the target task of face verification increased to $73.22\%$. We also performed empirical experiments
to study the effect of (a) descriptor size of the fine-tuned representations and (b) amount of
supervised training used during fine-tuning on the verification accuracies. We found that increasing
the dimensionality of the representations leads to better accuracies -- using a $1000-d$
fine-tuned descriptor, we were able to push the verification rates up to $74.13\%$. Similarly, using
larger amounts of supervised data also boosted performance. Our work is the first attempt at using
unsupervised learning methods in the limited, but nevertheless important domain of face images. We
believe that it brings forth newer opportunities to leverage the vast amounts of human-centric
multimedia data on the Internet for designing CNN based facial representations.

\section{Acknowledgements}
This work was partially supported by CEFIPRA, and by Research-I foundation, IIT Kanpur.

{\small
	\bibliographystyle{ieee}
	\bibliography{egbib}
}

\clearpage

\section*{\Large{Appendix}}
\input{appendix.tex}

\end{document}

%% file: appendix.tex
\subsection{Ablation Studies}
We also performed some ablation studies to understand the effect of different parameters of the fine-tuning process. In particular, we studied the effect of the size of the projection dimension, $p$ and the amount of supervised LFW data (as measured by the number of LFW face pairs used for fine-tuning) on the verification accuracies. We discuss some of the results in this section.
\vspace{0.8em}\\
\textbf{Projection Dimension.}
We ran experiments to see the effect of change in the projection dimension on the verification
accuracies. Specifically, we varied $p$ (with $d = 1024$) while learning the projection matrix $W \in \mathbb{R}^{p
	\times d}$ for the network trained on our face pairs dataset and having an input image size of $64
\times 64$. In Table \ref{tab:accuracy2}, the accuracies are shown to increase  with an increase in
the size of the projection dimension.

\begin{table}[H]
	\begin{center}
		\begin{tabularx}{0.42\textwidth}{@{}l*{2}{C}c@{}}
			\toprule
			\multicolumn{1}{c}{\multirow{2}{*}{Proj. Dim.}}  & \multicolumn{2}{c}{Input image size = $64 \times 64$} \\
			\multicolumn{1}{c}{}  &  Acc.  &  AUC \\
			\midrule
			\ \ \ $p = 128$   &  73.18  &  80.30  \\
			\ \ \ $p = 256$   & 73.22   &  80.57  \\
			\ \ \ $p = 512$   & 74.00   &  81.41  \\
			\ \ \ $p = 1000$   & 74.13   &  81.66  \\
			\bottomrule
		\end{tabularx}
		\caption{Effect of the dimensionality of the fine-tuned descriptors on verification accuracies.}
		\label{tab:accuracy2}
	\end{center}
	\vspace{-3em}
\end{table}

\textbf{Amount of Supervised Data.}
In Table \ref{tab:accuracy3}, we report verification accuracies by varying the amount of supervised
LFW face pairs data used for fine-tuning the unsupervised representations. The trends show that using larger amounts of training data improves performance.

\begin{table}[H]
	\begin{center}
		\begin{tabularx}{0.45\textwidth}{@{}l*{2}{C}c@{}}
			\toprule
			\multicolumn{1}{c}{\multirow{2}{*}{\# Face Pairs}}  & \multicolumn{2}{c}{Input image size = $64 \times 64$} \\
			\multicolumn{1}{c}{}  &  Acc.  &  AUC \\
			\midrule
			
			\ \ \ $1k$   & 70.89 & 78.62 \\
			\ \ \ $2k$   & 71.75 & 79.40 \\
			\ \ \ $5k$   & 73.49 & 80.38 \\
			\ \ \ $10k$  & 73.22 & 80.57 \\
			\ \ \ $20k$  & 73.85 & 81.13 \\
			
			\bottomrule
		\end{tabularx}
		\caption{Effect of the amount of supervised data during metric learning on the verification accuracies}
		\label{tab:accuracy3}
	\end{center}
	\vspace{-3em}
\end{table}

\subsection{ROC Curves}
\begin{figure}[H]
	\begin{center}
		\includegraphics[width=\columnwidth,trim=0 10 200 40, clip]{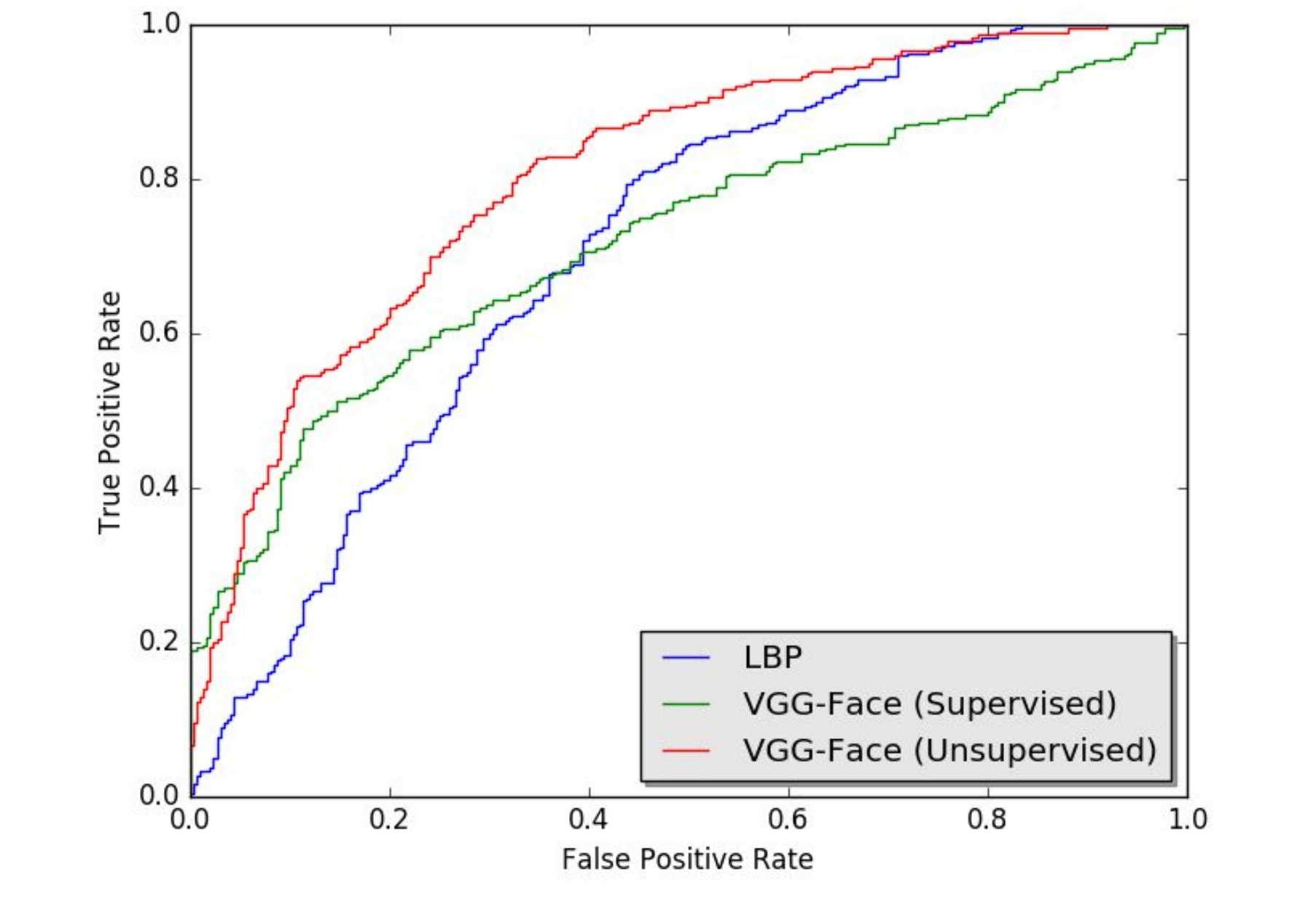}
		\vspace{-3em}
	\end{center}
	\caption{ROC Curves for verification on the LFW dataset}
	\label{fig:rocCurves}
	\vspace{-1.5em}
\end{figure}

We plot ROC curves for face verification on the LFW dataset.
We compare our unsupervised representations with two baselines -- LBP (hand-crafted features) and representations from the VGG-Face \cite{parkhi2015deep} network trained using supervised (identity labeled) dataset. Figure \ref{fig:rocCurves} shows the ROC curves for our (unsupervised) representations and the baselines.

